\documentclass[11pt,fleqn]{article}
\usepackage{amsfonts,ulem,url} 
\usepackage{amstext,verbatim,graphicx,ifthen,multirow,amsthm}

\usepackage{amssymb}
\usepackage{amsmath}
\usepackage{hhline}
\usepackage{graphicx}

\newcommand{\emse}{{\rm EMSE}}
\newcommand{\calt}{{\cal T}}
\newcommand{\est}{{\rm est}}
\newcommand{\calp}{{\Pi}}
\newcommand{\cals}{{\cal S}}

\newcommand{\mse}{{\rm MSE}}

\newcommand{\emset}{{\rm EMSE}_\tau}
\newcommand{\mset}{{\rm MSE}_\tau}

\newcommand{\msem}{{\rm MSE}_\mu}
 
\newcommand{\msemw}{{\rm MSE}_{\mu,W}}
\newcommand{\control}{{\rm control}}
\newcommand{\treat}{{\rm treat}}

\newcommand{\oy}{\overline{Y}}

\newcommand{\train}{{\rm tr}}

\newcommand{\test}{{\rm te}}

\newcommand{\obs}{{\rm obs}}

\newcommand{\tcv}{{\rm cv}}

\newtheorem{assumption}{Assumption}

\newcommand{\yin}{Y_i(0)}
\newcommand{\yie}{Y_i(1)}
\newcommand{\yio}{Y_i^\obs}
\newcommand{\mme}{\mathbb{E}}

\newcommand{\indep}{\perp\!\!\!\perp}

\def\monthname{\ifcase\month\or
  January\or February\or March\or April\or May\or June\or July\or
  August\or September\or October\or November\or December\fi}

\numberwithin{equation}{section}

\newcommand{\bge}{\begin{equation}}
\newcommand{\ene}{\end{equation}}

\newcommand{\mmv}{{\mathbb{V}}}

\def\monthname{\ifcase\month\or
January\or February\or March\or April\or May\or June\or
July\or August\or September\or October\or November\or December\fi}

\begin{document}

\title{Recursive Partitioning for Heterogeneous Causal Effects\thanks{%
{\small 
We are grateful for comments provided at seminars at the National Academy of Science Sackler Colloquium, the Southern Economics Association, the Stanford Conference on Causality in the Social Sciences, the MIT Conference in Digital Experimentation, Harvard, University of Washington, Cornell,  Microsoft Research, Facebook, KDD, the AAAI Embedded Machine Learning Conference, the University of Pennsylvania, the University of Arizona.  Part of this research was conducted while the authors were visiting Microsoft Research.
}}}
\author{
Susan Athey\thanks{{\small Graduate School of Business, Stanford University, and NBER. Electronic correspondence: athey@stanford.edu }}
\and
 Guido W. Imbens\thanks{{\small Graduate School of Business, Stanford University, and NBER. Electronic correspondence: imbens@stanford.edu }} 
}
\date{
First Draft: October 2013 \\ This Draft:
 \ifcase\month\or
January\or February\or March\or April\or May\or June\or
July\or August\or September\or October\or November\or December\fi \ \number%
\year
}

%\contributor{Submitted to Proceedings of the National Academy of Sciences of the United States of America}

\maketitle

\begin{abstract}
{In this paper we propose methods for estimating heterogeneity in causal effects in experimental and observational studies, and for conducting hypothesis tests about
the magnitude of the differences in treatment effects across subsets of the population.  We
provide a data-driven approach to partition the data into subpopulations which differ in the magnitude of their treatment effects.  The approach enables the construction of valid confidence intervals for treatment effects, even in samples with many covariates relative to the sample size, and without  ``sparsity'' assumptions.  To accomplish this, we propose an ``honest'' approach to estimation, whereby one sample is used to construct the partition
and another to estimate treatment effects for each subpopulation.   Our
approach builds on  regression tree methods, modified to optimize for goodness of fit in {\it treatment effects} and to account for honest
estimation.  Our model selection criteria focus on improving the prediction of treatment effects conditional on covariates, anticipating that bias will be eliminated by honest estimation, but also
accounting for the change in the variance of treatment effect estimates within each subpopulation as a result of the split.  We also address the challenge that the ``ground truth'' for a causal effect is not observed for any individual unit, so that standard approaches
to cross-validation must be modified. Through a simulation study, we show that honest estimation can result in substantial improvements in coverage of confidence intervals, where our method attains nominal coverage rates, without much sacrifice in terms of fitting
treatment effects.
}       
\end{abstract}

%\keywords{Potential Outcomes | Heterogeneous Treatment Effects | Causal Inference | Supervised Machine Learning | Cross-Validation}

%\abbreviations{LASSO-Least Absolute Shrinkage and Selection Operator; MSE-Mean Squared Error}

%\dropcap{I}n 
In this paper we study two closely related problems: first, estimating heterogeneity by covariates or features in causal effects in experimental or observational studies, and second, conducting inference about
the magnitude of the differences in treatment effects across subsets of the population.  Causal effects, in the Rubin Causal Model or potential outcome framework  we use here (\cite{R74},  \cite{H86}, \cite{IR15}), are comparisons between outcomes we observe and counterfactual  outcomes we would have observed under a different regime or treatment.  We introduce data-driven methods that
 select subpopulations to estimate more precisely average treatment effects and to test hypotheses about the
differences between the effects in different subpopulations.  For experiments, our method allows researchers to identify heterogeneity in treatment effects that was not specified in a 
pre-analysis plan, without concern about invalidating inference due to concerns about multiple testing.

Our approach is tailored for applications where there may be many attributes of a unit relative to the number of units observed, and where the functional form of the
relationship between treatment effects and the attributes of units is not known.  The supervised machine learning literature (e.g. \cite{HTF}) has developed a variety of effective methods for a closely
related problem, the problem of predicting outcomes as a function of covariates in similar environments.  The most popular approaches (e.g. regression trees (\cite{BFOS}), random forests (\cite{B01}), LASSO (\cite{T96}),  support vector machines (\cite{V98}), etc.) 
entail building a model of the relationship between attributes and outcomes, with a penalty parameter that penalizes model complexity. Cross-validation is often used to select the optimal level
of complexity (the one that maximizes predictive power without ``overfitting''). 

Within the prediction-based machine learning literature, regression trees differ from most other methods in that they produce a partition of the population according to covariates, whereby
all units in a partition receive the same prediction.  In this paper, we focus on the analogous goal of deriving a partition of the population according to
treatment effect heterogeneity, building on standard regression trees (\cite{BFOS}, \cite{B01}).  Whether the ultimate goal in an application is to derive a partition or fully personalized treatment effect estimates depends on the setting;
settings where partitions may be desirable include those where decision rules must be remembered, applied or interpreted by human beings or computers with limited processing power or memory. Examples include treatment guidelines to be used by physicians or even online personalization applications where 
having a simple lookup table reduces latency for the user. 
We show that an attractive feature of focusing on partitions is that we can achieve nominal converage of confidence intervals for estimated treatment effects
even in settings with a modest number of observations and many covariates. % (possibly even more covariates than observations).  Thus, 
Our approach has applicability even for settings such as clinical trials of drugs with only a few hundred patients, where the number of
patient characteristics is potentially quite large.  Our method may also be viewed as a complement to the use of ``pre-analysis plans''
where the researcher must commit in advance to the subgroups that will be considered for analysis.  It enables researchers to let
the data discover relevant subgroups without falling prey to concerns of multiple hypothesis testing that would invalidate p-values.   

A first challenge for our goal of finding a partition and then testing hypotheses about treatment effects is that many existing machine learning methods cannot be used directly for constructing confidence intervals. 
% In the standard prediction context, the predictions are not asymptotically normal and centered around the truth; many methods are asymptotically bias-dominated.  
This is because the methods are ``adaptive'': they use the training
data for model selection, so that spurious correlations between covariates and outcomes affect the selected model, leading to biases that disappear only slowly as the sample
size grows.  In some contexts, additional assumptions such as ``sparsity'' (only a few covariates affect the outcomes) can be applied to guarantee
consistency or asymptotic normality of predictions (\cite{WA}).  In this paper, we use an alternative approach that places no
restrictions on model complexity, which we refer to as ``honesty.'' We say that a model is ``honest'' if it does
not use the same information for selecting the model structure (in our case, the partition of the covariate space) and for estimation  given a model structure. 
We accomplish this by splitting the training sample into two parts, one for constructing the tree (including the cross-validation step), and a second for estimating treatment
effects within leaves of the tree, implying that %.  One attractive feature of this approach is that 
the asymptotic properties of treatment effect estimates within the partitions are the same
as if the partition had been exogenously given.  Although there is a loss of precision due to sample splitting (which reduces sample size in each step of estimation), there is a benefit for fit in terms of eliminating bias that offsets at least part of the cost.

A novel contribution of this paper is to show that criteria for both constructing the partition
% (which boils down to ``splitting'' criteria for leaves of the tree) 
and cross-validation change when we anticipate honest estimation.  In the first stage of estimation, the criteria is the expectation of
the mean-squared error when treatment effects are re-estimated in the second stage.  Crucially, we anticipate that second-stage estimates of treatment effects will be unbiased in each leaf,
since they will be performed on a separate (and independent) sample.  In that case, splitting and cross-validation criteria are adjusted to ignore systematic bias in estimation, and focus 
instead on the tradeoff between more tailored prediction (smaller leaf size) and the variance that will arise in the second (honest estimation) stage due to noisy estimation within small leaves. 

A second and perhaps more fundamental challenge to applying machine learning methods such as regression trees \cite{BFOS} off-the-shelf to the problem of causal inference is that regularization approaches based on cross-validation
typically rely on observing the ``ground truth,'' that is, actual outcomes in a cross-validation sample.  However, if our goal is to minimize the mean-squared error of treatment
effects, we encounter what \cite{H86} calls the ``fundamental problem of causal inference'': the causal effect is not observed for any individual unit, and so we don't directly
have a ground truth.  We address this  by proposing approaches for constructing unbiased estimates of the mean-squared error of the causal effect of the treatment. 

%In the paper, we contrast our approach with several that have been raised in the existing literature, including using statistical criteria based on model fit or statistical tests for reatment effect hetergoeneity.  We show that with some approximations, our criteriacan be expressed as a way of balancing test statistics based on model fit or treatmenteffect heterogeneity; we showtheoretically as well as through simulations that in general our criteria do a better job selecting models that minimize mean-quared error of treatment effects, and our approach gets nominal coverage rates for confidenceintervals, unlike the existingapproaches. 

Using theoretical arguments and  a simulation exercise we evaluate the costs and benefits of honest estimation and compare our approach with previously proposed ones.
We find in the settings we consider that 
honest estimation results in improvements in fit, in some cases very large improvements, over a more traditional ``adaptive'' estimation approach.

% In our working paper \cite{AI}, we apply the method to an application, a large-scale field experiment re-ranking results on a search engine.    
% Although we focus in the current paper mostly on regression tree methods, the approach extends directly to other popular supervised machine learning methods. 

\section{The Problem}\label{section:setup}

\subsection{The Set Up}

We consider a setup where there are $N$ units, indexed by $i=1,\ldots,N$.   
We postulate the existence of a pair of potential outcomes for each unit, $(\yin,\yie)$ (following the potential outcome or Rubin Causal Model \cite{R74}, \cite{H86}, \cite{IR15}, with the unit-level causal effect defined as the difference in potential outcomes, $ \tau_i=\yie-\yin.$
Let $W_i\in\{0,1\}$ be the binary indicator for the treatment, with $W_i=0$ indicating that unit $i$ received the control treatment, and $W_i=1$ indicating that unit $i$ received the active treatment. 
 The realized outcome for unit $i$ is the potential outcome corresponding to the treatment received:
\[ \yio=Y_i(W_i)=\left\{\begin{array}{ll} 
\yin\hskip1cm & {\rm if}\ W_i=0,\\
\yie\hskip1cm & {\rm if}\ W_i=1.\end{array}\right.
\]
Let $X_i$ be a  $K$-component vector of features, covariates or pretreatment variables, known not to be affected by the treatment. 
Our data consist of the triple $(\yio,W_i,X_i)$, for $i=1,\ldots,N$, which are regarded as an i.i.d sample drawn from a large population. Expectations and probabilities will refer to the distribution induced by the random sampling, or by the (conditional) random assignment of the treatment.
We assume that observations are exchangeable, and that there is no interference (the stable unit treatment value assumption, or sutva \cite{R78}). This assumption may be violated in settings where some units are connected through networks.
Let $p={\rm pr}(W_i=1)$ be the marginal treatment probability, and let $e(x)={\rm pr}(W_i=1|X_i=x)$ be the conditional treatment probability (the ``propensity score" as defined by \cite{RR}). In a randomized experiment with constant treatment assignment probabilities $e(x)=p$ for all values of $x$.

\subsection{Unconfoundedness}

Throughout the paper, we maintain the assumption of randomization conditional on the covariates, or ``unconfoundedness'' (\cite{RR}), formalized as:
\begin{assumption}\label{unconfoundedness}{\sc (Unconfoundedness)}
\[W_i\ \indep\ \Bigl( Y_i(0),Y_i(1)\Bigr)\ \Bigl|\ X_i.\]\end{assumption}
This assumption, sometimes referred to as ``selection on observables'' in the econometrics literature, is satisfied in a randomized experiment without conditioning on covariates, but also may be
justified in observational studies if the researcher is able to observe all the variables that affect the unit's receipt of treatment and are associated with the potential outcomes. 

To simplify exposition, in the main body of the paper we maintain the stronger assumption of {\it complete randomization,} whereby $W_i \indep (Y_i(0), Y_i(1),X_i)$.  Later we show that by using propensity score weighting \cite{R74}, we can  adapt all of the methods to that case.  

\subsection{Conditional Average Treatment Effects and Partitioning}

Define the { conditional average treatment effect} (CATE)
\[ \tau(x) \equiv
\mme[\yie-\yin|X_i=x].\]
%and the population average treatment effect\[ \tau=\mme[Y_i(1)-Y_i(0)]=\mme[\tau(X_i)].\]
A large part of the causal inference literature (e.g. \cite{IR15}, \cite{P00}) is focused on estimating the population (marginal) average treatment effect $\mme[Y_i(1)-Y_i(0)]$.
The main focus of the current paper is on obtaining accurate estimates of and inferences for the conditional average treatment effect $\tau(x)$. 
We are interested in estimators $\hat\tau(\cdot)$ that %satisfy certain properties. First, we focus on estimators that 
are based on partitioning the feature space, and do not vary within the  partitions.

\section{Honest Inference for Population Averages}

Our approach departs from conventional classification and regression trees (CART) in two fundamental ways. First, we focus on estimating conditional average treatment effects rather than predicting
outcomes. 
This creates complications for conventional methods because we do not observe unit level causal effects for any unit.
Second, we impose a separation between constructing the partition and estimating effects within leaves of the partition, using separate samples for the two tasks,
in what we refer to as ``honest'' estimation.  We contrast `'honest'' estimation with ``adaptive'' estimation used in conventional CART, where the same data is used
to build the partition and estimate leaf effects.  In this section we introduce the changes induced by honest estimation in the context of the conventional prediction setting; in the next section we consider causal effects.  In the discussion in this section we observe for each unit $i$ a pair of variables $(Y_i,X_i)$, with the interest in the conditional expectation $\mu(x) \equiv \mme[Y_i|X_i=x]$.

\subsection{Set Up}
We begin by defining key concepts and functions. First, a tree or partitioning $\calp$ corresponds to a partitioning of the feature space $\mathbb{X}$, with $\#(\calp)$ the number of elements in the partition. We write 
\[ \calp=\{\ell_1,\ldots,\ell_{\#(\calt)}\},\ \ {\rm with}\ 
\cup_{j=1}^{\#(\calp)} \ell_j=\mathbb{X}.\] Let $\mathbb{P}$ denote the space of partitions.
Let $\ell(x;\calp)$ denote the leaf $\ell\in\calp$ such that $x\in\ell$.
Let $\mathbb{S}$ be the space of data samples from a population. Let $\pi:\mathbb{S}\mapsto\mathbb{P}$ be an algorithm that on the basis of a sample $\cals \in \mathbb{S}$ constructs a  partition.
As a very simple example, suppose the feature space is $\mathbb{X}=\{L,R\}$. In this case there are two possible partitions,
$\Pi_N=\{L,R\}$ (no split), or $\Pi_S=\{\{L\},\{R\}\}$ (full split), and so the space of trees is $\mathbb{P}=\{\Pi_N,\Pi_S\}=\{\{L,R\},\{\{L\},\{R,\}\}\}$.
Given a sample $\cals$, the average outcomes in the two subsamples are $\oy_L$ and $\oy_R$.  A simple example of an algorithm is one that splits if the difference in average outcomes exceeds a threshold $c$:
\[ \pi(\cals)=
\left\{
\begin{array}{ll}
\{\{L,R\}\} \hskip1cm & {\rm if}\  \oy_L-\oy_R\leq c,\\
\{\{L\},\{R\}\} & {\rm if}\  \oy_L-\oy_R>c.\end{array}\right.
\]
\noindent The potential bias in leaf estimates from adaptive estimation can be seen in this simple example.  While $\oy_L-\oy_R$ is in general an unbiased estimator
for the difference in the population conditional means $\mu(L)-\mu(R)$, if we condition on finding that
$\oy_L-\oy_R\geq c$ in a particular sample, we expect that $\oy_L-\oy_R$ is larger than the population analog.

Given a partition $\calp$, 
define the conditional mean function $\mu(x;\calp)$ as
\[ \mu(x;\calp) \equiv \mme[Y_i|X_i\in\ell(x;\calp)]=\mme[\mu(X_i)|X_i\in\ell(x;\calp)],\]
which can be viewed as a step-function approximation to $\mu(x)$.
Given a sample $\cals$ the estimated counterpart is
%define an estimated conditional mean at $x\in\mathbb{X}$ as the average outcome for units in the same leaf as $x$:
\[ \hat\mu(x;\cals,\calp) \equiv \frac{1}{\#(i\in\cals: X_i\in\ell(x;\calp))}\sum_{i\in\cals:X_i\in\ell(x;\calp)} Y_i,\]
which is unbiased for $\mu(x;\calp)$.
We index this estimator by the sample because we need to be precise which sample is used for estimation of the regression function.

\subsection{The Honest Target} A central concern in this paper is the criterion used to compare alternative estimators; following
much of the literature, we focus on Mean-squared error (MSE) criteria, but we will modify these criteria in a variety of ways.
%captures the goodness of fit of an estimator, and this is the criterion we will  focus on.
  For the prediction case, we adjust the MSE by $\mme[Y^2_i]$; since this does not depend on an estimator, subtracting it does not affect how the criterion ranks estimators.  
Given a partition $\Pi$, define the mean squared error, where we average over a test sample $\cals^\test$ and the conditional mean is estimated on an estimation sample $\cals^\est$, as
\[  \mse(\cals^\test,\cals^\est,\calp) \equiv %\]\[
\frac{1}{\#(\cals^\test)}\sum_{i\in\cals^\test}
\left\{\left(Y_i-\hat\mu(X_i;\cals^\est,\calp)\right)^2-Y^2_i \right\}.\]
The (adjusted) expected mean squared error is the expectation of $ \mse(\cals^\test,\cals^\est,\calp)$ over the test sample and the estimation sample:
\[ \emse(\calp) \equiv \mme_{\cals^\test,\cals^\est}\left[ \mse(\cals^\test,\cals^\est,\calp)\right],\]
where the test and estimation samples are independent.
 In the algorithms we consider, we will consider a variety of estimators for the (adjusted) EMSE,
all of which take the form of MSE estimators $\mse(\cals^\test,\cals^\est,\calp)$,
evaluated at the units in sample $\cals^\test$, with the estimates based on sample $\cals^\est$ and the tree $\calp$.  For brevity in this paper we will henceforth omit the term ``adjusted'' and
abuse terminology slightly by referring
to these objects as MSE functions.  

Our ultimate goal is to construct and assess algorithms $\pi(\cdot)$ that maximize  the ``honest'' criterion
\[ Q^H(\pi)
\equiv 
 -\mme_{\cals^\est,\cals^\est,\cals^\train}\left[\mse(\cals^\test,\cals^\est,\pi(\cals^\train))\right]. \]
Note that throughout the paper we focus on maximixing criterion functions, which typically involve the negative of mean-squared-error expressions.

\subsection{The Adaptive Target}
%onventional Approach to Splitting and Crossvalidation}

In the conventional CART approach the target is slightly different:
\[Q^{C}(\pi) \equiv-\mme_{\cals^\test,\cals^\train}\left[\mse(\cals^\test,\cals^\train,\pi(\cals^\train))\right],\]
where the same training sample is used to construct and estimate the tree. Compared to our target $Q^H(\pi)$ the difference is that in our approach  different samples $\cals^\train$ and $\cals^\est$ are used for construction of the tree and estimation of the conditional means respectively.
%To contrast our criterion with the criterion used in the conventional approach of CART, the criterion our honest approach uses is $Q_Y^{H}(\pi(\cals^{tr}))$ rather than $ -\emse(\cals^{tr},\pi(\cals^{tr}))$, which is the conventional target. There are two key 
%The key difference between the two approaches is that we use $\cals^E$ instead of $\cals^\train$ to estimate leaf means, and second (and closely related), $\cals^E$ is unknown at the time the target is evaluated, while $\cals^\train$ is known, and thus the leaf estimates are known in the conventional criterion.
% Indeed, the conventional approach does not distinguish the two components of tree estimation (partition construction and leaf estimation). 
 We refer to the conventional CART approach as ``adaptive,'' and our approach as ``honest.''
% with the key distinction being whether the same data is used to build trees and estimate leaf effects, or not.  

In practice there will be costs and
benefits of the honest approach relative to the adaptive approach. The cost is sample size; given a data set, putting some data in the estimation sample leaves fewer units for the training data set.
%, we may put 45\% of the data in the training set, 45\% in the estimation sample, and 10\% in a test (hold-out) sample for evaluating the predictions at the end.  In contrast, for adaptive estimation, 90\% of the data may be used in the training set.
%  Leaving aside for the moment the sample size issue, imagine that the sample $\cals^{\train}$ is fixed, and that we have available a second independent sample $\cals^{\est}$ of the same size.  Then,
The advantage of honest
estimation is that it avoids a  problem of adaptive estimation, which is that spurious extreme values of $Y_i$ are likely to be placed into the same leaf as other extreme
values by the algorithm $\pi(\cdot)$,
and thus the sample means (in sample $\cals^{\train}$) of the elements of $\pi(\cals^\train)$ are more extreme than they would be in an independent sample. 
% As a result, for a fixed algorithm $\pi$, $Q^C(\pi)>Q^H(\pi)$.   %, so that if the estimation and training sample are the same size ($N^\est=N^\train$), then
%\[ Q^{C}(\pi)<Q^{H}(\pi).\]
%\[ \mme_{\cals^{\test},\cals^{\est},\cals^\train} \left[\mse(\cals^{\test},\cals^{\est},\pi(\cals^{\train}))\right]<\]
%\[\mme_{\cals^{\test}, \cals^\train} \left[ \mse(\cals^{\test},\cals^{\train},\pi(\cals^{\train}))\right].\]

\subsection{The Implementation of CART}

There are two distinct parts of the conventional CART algorithm, initial
tree building and cross-validation to select a complexity parameter used for pruing. Each part of the algorithm relies on a criterion
function based on mean-squared error.  In this paper we will take as given the overall structure of the CART algorithm (e.g., \cite{BFOS}, \cite{HTF}), and our
focus will be on modifying the criteria.  %We begin by introducing the details of the conventional criteria.  
 
In the tree-building phase, 
CART recursively partitions the observations of the training sample.  For each leaf, the algorithm evaluates all candidate splits of that leaf (which induce alternative partitions $\Pi$) 
using a ``splitting'' criterion that we refer to as the ``in-sample'' goodness of fit criterion $-\mse(\cals^{\train},\cals^{\train},\calp)$.
%For the class of constaint-within-leaf estimators $\hat\mu(x;\cals,\calp)$, it can be shown that when the estimation and test sample are identical, then\[ -\mse(\cals,\cals,\calp)=\frac{1}{\#(\cals)}\sum_{i\in\cals} \hat\mu^2(X_i;\cals,\calp).\]
It is well-understood that the conventional criterion leads to ``over-fitting,'' a problem that is solved by cross-validation to select a penalty on tree depth. 
%To help set the stage for our modifications to CART, we discuss the over-fitting problem in the context of our notation. 
The in-sample goodness of fit criterion will always improve with additional splits, even though additional refinements of a partition $\calp$ might in fact increase 
the expected mean squared error, especially when the leaf sizes become small.
%This discrepancy arises  mainly because 
The reason is that
%for two reasons.  First, 
the criterion ignores the fact that smaller leaves lead to higher-variance estimates of leaf means.
% Our honest criterion $\emse(\calp)$ incorporates this variance by explicitly taking into account the variation induced by estimating the leaves; below, we propose an estimator that incorporates this effect.  Second, as discussed above, adaptive estimation leads leaf estimates to be more extreme in the training sample than they will be in an independent test sample, so that the ``in-sample'' goodness of fit under-estimates what the  goodness of fit will be on an independent test sample.

To account for this factor, the conventional approach to
avoiding ``overfitting'' is to add a penalty term to the criterion that is equal to a constant  times the number of splits, so that essentially we only consider splits where the improvement in a goodness-of-fit criterion is above some threshold.
The penalty term is choosen to maximize a goodness of fit criterion in cross-validation samples. 
In the conventional cross-validation the training sample is repeatedly split into two subsamples, the $\cals^{\train,\train}$ 
sample that is used to build a new tree as well as estimate the conditional means and the $\cals^{\train,\tcv}$ sample that is used to evaluate the estimates.  
We ``prune'' the tree using a penalty parameter  that represents the cost of a leaf. 
We choose the optimal penalty parameter by evaluating the trees associated with each value of the penalty parameter.  The goodness of fit criterion for cross-validation can be written as
%\[-\frac{1}{N^{\train,\tcv}}\sum_{i\in\cals^{\train,\tcv}}\left\{Y_i-\hat\mu(X_i;\cals^{\train,\train},\calp)\\right\}^2+Y^2_i\]
$
- \mse(\cals^{\train,\tcv},\cals^{\train,\train},\calp).$
Note that the cross-validation
criterion directly addresses the issue we highlighted with the in-sample goodness of fit criterion, since $\cals^{\train,\tcv}$ is independent of $\cals^{\train,\train}$,
and thus too-extreme estimates of leaf means will be penalized.  The issue
that smaller leaves lead to noisier estimates of leaf means is implicitly incorporated by the fact that a smaller leaf penalty will lead to 
deeper trees and thus smaller leaves, and the noisier estimates will lead to larger average $\mse$ across the cross-validation samples. 

\subsection{Honest Splitting}

In our honest estimation algorithm, we modify CART in two ways.  First, we use an independent sample $\cals^\est$ instead of $\cals^\train$ to estimate leaf means.
Second (and closely related), we modify our splitting and cross-validation criteria to incorporate the fact that we will generate unbiased estimates using $\cals^\est$ for
leaf estimation (eliminating one aspect of over-fitting), where $\cals^\est$ is treated as a random variable in the tree building phase. In addition, we explicitly incorporate the fact that finer partitions generate greater variance in leaf estimates. 
% In contrast, the conventional approach does not distinguish the two components of tree estimation (partition construction and leaf estimation).    

To begin developing our criteria, let us
expand $\emse (\calp)$:
\[
-\emse (\calp) =
%-\mme_{\cals^\test,\cals^\est}[\mse(\cals^\test,\cals^\est,\calp)]=\]\[
-\mme_{ (Y_i,X_i),\cals^\est}[(Y_i - \mu(X_i;\calp))^2-Y_i^2]\]
\[-\mme_{X_i,\cals^\est}\left[\left(\hat\mu(X_i;\cals^\est,\calp)-\mu(X_i;\calp)\right)^2\right]=\]
\[
\mme_{X_i}[\mu^2(X_i;\calp)]-\mme_{\cals^\est, X_i}\left[\mmv(\hat\mu^2(X_i;\cals^\est,\calp)\right],\]
where we exploit the equality $\mme_\cals[\hat\mu(x;\cals,\calp)]=\mu(x;\calp)$.

We wish to estimate $-\emse (\calp)$ on the basis of the training sample $\cals^\train$ and knowledge of the sample size of the estimation sample $N^\est$.
To construct an estimator for the second term, observe that within each leaf of the tree there is an unbiased estimator for the variance of the estimated mean in that leaf. 
Specifically, to estimate the variance of $\hat\mu(x;\cals^\est,\calp)$ on the training sample we can use
\[ \widehat{\mmv}(\hat\mu(x;\cals^\est,\calp)) \equiv \frac{S^2_{\cals^\train}(\ell(x;\calp))}{N^\est(\ell(x;\calp))},\]
where $ S^2_{\cals^\train}(\ell)$ is the within-leaf variance, 
to estimate the variance.
%The within-leaf variance estimator, based on the training sample, is
%\[ S^2_{\cals^\train}(\ell) \equiv \frac{1}{\#(\{i\in\cals^\train:X_i\in \ell\})}\sum_{i\in\cals^\train:X_i\in\ell} (Y_i-\hat\mu(X_i;\cals^\train,\calp))^2.\]
We then weight this by the leaf shares $p_\ell$ to estimate the expected variance.
%\[ \widehat{\mme}\left[\mmv(\hat\mu^2(X_i;\cals^\est,\calp)|i\in\cals^\test\right]=\sum_{\ell\in\calp} p_\ell\cdot \frac{S^2_{\cals^\train}(\ell)}{N^\est(\ell)}.\]
Assuming the leaf shares are approximately the same in the estimation and training sample, we can approximate this variance estimator by
\[ \widehat{\mme}\left[\mmv(\hat\mu^2(X_i;\cals^\est,\calp)|i\in\cals^\test\right]
\equiv \frac{1}{N^\est}\cdot \sum_{\ell\in\calp} S^2_{\cals^\train}(\ell).\]

To estimate the average of the squared outcome $\mu^2(x;\calp)$ (the first term of the target criterion), 
we can use the square of the estimated means in the training sample $\hat\mu^2(x;\calp)$, minus an estimate of its variance,
\[
\widehat{\mme}[\mu^2(x;\calp)]=\hat\mu^2(x;\cals^\train,\calp)-
\frac{S^2_{\cals^\train}(\ell(x;\calp))}{N^\train(\ell(x;\calp))}.
\]
Combining these estimators leads to the following unbiased estimator for $\emse(\calp)$, denoted $\widehat{\emse}(\cals^\train,N^\est,\calp)$:
\[\frac{1}{N^\train}\sum_{i\in\cals^\train} \hat\mu^2(X_i;\cals^\train,\calp)-
\left(\frac{1}{N^\train}+\frac{1}{N^\est}\right)\cdot \sum_{\ell\in\calp} S^2_{\cals^\train}(\ell)
.\]
In practice we use the same sample size for the estimation sample and the training sample, so we use as the estimator
\[
\widehat{\emse}(\cals^\train,\calp) \equiv % \]\[
\frac{1}{N^\train}\sum_{i\in\cals^\train} \hat\mu^2(X_i;\cals^\train,\calp)-
\frac{2}{N^\train}\cdot \sum_{\ell\in\calp} S^2_{\cals^\train}(\ell)
.\]
Comparing this to the criterion used in the conventional CART algorithm, which can be written as
\[\mse(\cals^\train,\cals^\train,\calp)=\frac{1}{N^\train}\sum_{i\in\cals^\train} \hat\mu^2(X_i;\cals^\train,\calp),\]
  the difference comes from the terms involving the variance. In the prediction setting the adjustment makes very little difference. Because of the form of the within-leaf sample variances, it follows that the gain from a particular split according to the unadjusted criterion 
$\mse(\cals^\train,\cals^\train,\calp)$ is proportional to the gain based on $ \widehat{\emse}(\cals^\train,\calp)$, with the constant of proportionality a function of the leaf size. Thus, in contrast to the treatment effect case discussed below, the variance adjustment does matter much here.

%The introduction of the variance term implies that the criterion does not always improve with additional splits; the criterion anticipates that splits will increase the variance of the leaf estimates obtained fom the estimation sample $\cals^\est$.

%ADD TWO-THREE SENTENCES OR EQUATION TO DISCUSS RELATIONSHIP BETWEEN S2 and MSE.  DISCUSS THIS.

\subsection{Honest Crossvalidation}

Even though $\widehat{\emse}(\cals^\train,\calp)$ is approximately  unbiased as an estimator of our ideal criterion $\emse(\calp)$ for a fixed $\calp$, 
it is not unbiased when we use it repeatedly to evaluate splits using recursive partitioning on the training data $\cals^\train$.  The reason is that 
initial splits tend to group together observations with similar, extreme outcomes. So after the training data has been divided once, the sample variance of observations
in the training data within a given leaf is on average lower than the sample variance would be in a new, independent sample.  %Similarly, expected outcomes are on average more extreme.  
Thus, $\widehat{\emse}(\cals^\train,\calp)$ is likely to overstate goodness of fit as we grow a deeper and deeper tree,
implying that cross-validation  can still play an important role with our honest estimation approach, 
though perhaps less so than in the conventional CART.

%Now consider the criterion for cross-validation. 
 Because the conventional CART cross-validation criterion 
does not account for honest estimation %and the extent to which it eliminates the problem of too-extreme estimates of leaf means from the adaptive approach
we consider the analogue of our unbiased estimate of the criterion, 
which accounts for honest estimation by evaluating a partition $\calp$ 
using only outcomes for  units from the cross-validation sample $\cals^{\train,\tcv}$:
\[-
\widehat{\emse}(\cals^{\train,\tcv},\calp)
%\hat{Q}_Y^{cv,H}(\cals^\train,N^{\est}_{\Pi},\calp) \equiv 
\]
%\[=\frac{1}{N^{\train,\tcv}}\sum_{i\in\cals^{\train,\tcv}} \hat\mu^2(X_i;\cals^{\train,\tcv},\calp)- \frac{2}{N^{\train,\tcv}} \sum_{\ell\in\pi(\cals^{\train,\train})} S^2_{\cals^{\train,\tcv}}(\ell) . \]
This estimator for the honest criterion is unbiased, although it may have higher variance than 
$ \mse(\cals^{\train,\tcv},\cals^{\train,\train},\calp)$ due to the small sample
size of the cross-validation sample.

\section{Honest Inference for Treatment Effects}

In this section we change the focus to estimating conditional average treatment effects instead of estimating conditional population means.  We refer to the estimators developed in this section as ``Causal Tree'' (CT) estimators.
%We discuss our approach to honest splitting and cross-validation. 
The setting with treatment effects creates some specific problems because we do not observe the value of the treatment effect whose conditional mean we wish to estimate. This complicates the calculation of the criteria we introduced in the previous section. However, a key point of this paper is that we can estimate these criteria and use those estimates for splitting and cross-validation.

We now observe in each sample the
triple $(Y^\obs_i,X_i,W_i)$. 
%The fact that we do not observe the unit-level causal effects $\tau_i=Y_i(1)-Y_i(0)$ creates a number of complications.
For a sample $\cals$ let $\cals_\treat$ and $\cals_\control$ denote the subsamples of treated and control units respectively, with cardinality $N_\treat$ and $N_\control$ respectively, and let $p=N_\treat/N$ be the share of treated units.
The concept of a tree remains the same as in the previous section. Given a tree $\Pi$, define for all $x$ and both treatment levels $w$
the population average outcome
\[\mu(w,x;\calp) \equiv \mme\left[\left. Y_i(w) \right| X_i\in\ell(x;\calp) \right], \]
and the 
average causal effect 
\[\tau(x;\Pi) \equiv \mme\left[\left. Y_i(1)-Y_i(0) \right| X_i\in\ell(x;\calp) \right]. \]
The estimated counter parts are
\[ \hat\mu(w,x;\cals,\calp) \equiv
\frac{1}{\#(\{i\in\cals_w:X_i\in\ell(x;\calp)\})}\sum_{i\in\cals_w:X_i\in\ell(x;\calp)} Y_i^\obs,\]
and\[ \hat\tau(x;\cals,\calp) \equiv \hat\mu(1,x;\cals,\calp)- \hat\mu(0,x;\cals,\calp).\]
Define the mean-squared error for treatment effects as
\[ 
\mset(\cals^\test,\cals^\est,\calp) \equiv %\]\[
\frac{1}{\#(\cals^\test)}\sum_{i\in\cals^\test}
\left\{\left(\tau_i-\hat\tau(X_i;\cals^\est,\calp)\right)^2-\tau^2_i \right\},\]
and define $\emset(\calp)$ to be its expectation over the estimation and test samples,
\[ \emset(\calp) \equiv \mme_{\cals^\test,\cals^\est}\left[\mset(\cals^\test,\cals^\est,\calp)\right].\]

A key challenge %, and one that has not to our knowledge received attention in the literature to date, 
is
that the workhorse mean-squared error function $\mset(\cals^\test,\cals^\est,\calp)$ is {\it infeasible},
because we do not observe the $\tau_i$.
However, we show below that we can estimate it.

\subsection{Modifying Conventional CART for Treatment Effects}

Consider first modifying conventional (adaptive) CART to estimate heterogeneous treatment effects.  Note that in the prediction case, using the fact that $\hat\mu$ is constant within each leaf, we can write
\[ \msem(\cals^\test,\cals^\train,\calp)=-\frac{2}{N^\train}\sum_{i\in\cals^\test} \hat\mu(X_i;\cals^\test,\calp)
\cdot \hat\mu(X_i;\cals^\train,\calp)
\]
\[+\frac{1}{N^\train}\sum_{i\in\cals^\test} \hat\mu^2(X_i;\cals^\train,\calp).\]
In the treatment effect case we can use the fact that 
\[ \mme_{\cals^\test}\left[\tau_i|i \in \cals^\test: i \in \ell(x,\calp) \right] = \mme_{\cals^\test}\left[\hat\tau(x;\cals^\test,\calp)\right] \]
to construct an unbiased estimator of $\mset(\cals^\test,\cals^\train,\calp)$: 
\[\widehat{\mse}_\tau(\cals^\test,\cals^\train,\calp) \equiv -\frac{2}{N^\train}\sum_{i\in\cals^\test} \hat\tau(X_i;\cals^\test,\calp)
\cdot \hat\tau(X_i;\cals^\train,\calp)
\]
\[+\frac{1}{N^\train}\sum_{i\in\cals^\test} \hat\tau^2(X_i;\cals^\train,\calp).\]
This leads us to propose, by analogy to CART's in-sample mean-squared error criterion $-\msem(\cals^\train,\cals^\train,\calp)$,
\[ -\widehat{\mse}_\tau(\cals^\train,\cals^\train,\calp) = \frac{1}{N^\train}\sum_{i\in\cals^\train} \hat\tau^2(X_i;\cals^\train,\calp),\]
as an estimator for the infeasible
in-sample goodness of fit criterion.

For cross-validation we used in the prediction case $-{\mse}_\mu(\cals^{\train,\tcv},\cals^{\train,\train},\calp)$. Again, the treatment
effect analog is infeasible, but we can use an unbiased estimate of it, which leads to
$-\widehat{\mse}_\tau(\cals^{\train,\tcv},\cals^{\train,\train},\calp)
.$

\subsection{Modifying the Honest Approach}
The honest approach described in the previous section for prediction problems also needs to be modified for the treatment effect setting. Using the same expansion as before, now applied to the treatment effect setting, we find
\[
-\emse_\tau (\calp)=
\mme_{X_i}[\tau^2(X_i;\calp)]-\mme_{\cals^\est, X_i}\left[\mmv(\hat\tau^2(X_i;\cals^\est,\calp)\right].\]
For splitting we can estimate both components of this expectation using only the training sample. This leads to an estimator for the infeasible criterion that depends only on $\cals^\train$:
\[
\widehat{\emse}_\tau(\cals^\train,\calp) \equiv %\]\[
\frac{1}{N^\train}\sum_{i\in\cals^\train} \hat\tau^2(X_i;\cals^\train,\calp)
\]
\[-
\frac{2}{N^\train}\cdot \sum_{\ell\in\calp}
\left( \frac{S^2_{\cals^\train_\treat}(\ell)}{p}+ \frac{S^2_{\cals^\train_\control}(\ell)}{1-p}\right)
.\]
For cross-validation we use the same expression, now with the cross-validation sample:
$\widehat{\emse}_\tau(\cals^{\train,\tcv},\calp)$.
% = %\]\[
%\frac{1}{N^{\train,\tcv}}\sum_{i\in\cals^{\train,\tcv}} \hat\tau^2(X_i;\cals^{\train,\tcv},\calp)
%\]
%\[-\frac{2}{N^{\train,\tcv}}\cdot \sum_{\ell\in\calp}\left( \frac{S^2_{\cals^{\train,\tcv}_\treat}(\ell)}{p}+ \frac{S^2_{\cals^{\train,\tcv}_\control}(\ell)}{1-p}\right).\]

These expressions are directly analogous to the criteria we proposed for the honest version of CART in the prediction case.  The criteria reward a partition for finding strong heterogeneity in treatment effects, and penalize a partition
that creates variance in leaf estimates.  One difference with the prediction case,
however, is that in the prediction case, the two terms are proportional; whereas for the treatment effect case they are not. It is possible to reduce the variance of a treatment effect estimator by introducing a split,
even if both child leaves have the same average treatment effect, if a covariate affects the mean outcome but not treatment
effects. In such a case, the split results in more homogenous leaves, and thus lower-variance estimates of the means of the
treatment group and control group outcomes. Thus, the distinction
between adaptive and honest splitting criterion will be more pronounced in this case. 

The cross-validation criterion estimates treatment effects within
leaves using the $\cals^{\train,\tcv}$ sample rather than $\cals^{\train,\train}$, to account for the fact that leaf estimates will subsequently be constructed using an estimation sample
that is independent of the training sample.

\section{Four  Partitioning Estimators for Causal Effects}

In this section we briefly summarize our CT estimator, and then describe three alternative types of estimators.  We compare CT to the alternatives theoretically and through simulations. For each of the four types there is an adaptive version and an honest version, where the latter takes into account that estimation will be done on a sample separate from the sample used for constructing the partition, leading to a total of eight estimators. Note that further variations
are possible; for example, one could use adaptive splitting and cross-validation methods to construct a tree, but still perform honest estimation on a separate sample.  We do not consider those variations in this paper.

\subsection{Causal Trees (CT)} 

The discussion above developed our preferred estimator, Causal Trees. To summarize, for the adaptive version of causal trees, denoted CT-A, we use for splitting the objective $-\widehat{\mse}(\cals^\train,\cals^\train,\calp)$. For cross-validation we use the same objective function, but evaluated at the samples $\cals^{\train,\tcv}$ and $\cals^{\train,\train}$, namely
$-\widehat{\mse}(\cals^{\train,\tcv},\cals^{\train,\train},\calp)$.
For the honest version, CT-H, the splitting objective function is $-\widehat{\emse}(\cals^\train,\calp)$. For cross-validation we use the same objective function, but now evaluated at the cross validation sample, 
$-\widehat{\emse}(\cals^{\train,\tcv},\calp)$.

\subsection{Transformed Outcome Trees (TOT)}

Our first alternative method is based on the insight that by using a transformed version of the outcome $Y^*_i=(Y_i-W_i)/(p\cdot (1-p))$, it is possible to 
use off-the-shelf regression tree methods to focus splitting and cross-validation on treatment effects rather than outcomes.  Similar approaches are 
 used in   \cite{BL},  \cite{DLL},  \cite{S07},  and \cite{WP}.   Because $\mme[Y^*_i|X_i=x]=\tau(x)$, off-the-shelf CART methods can be used directly, where estimates of the sample average of $Y^*_i$ within each leaf
can be interpreted as estimates of treatment effects. This ease of application is the key attraction of this method. The main drawback
(relative to CT-A) is that in general it is not efficient because it does not use the information in the treatment indicator beyond the construction of the transformed outcome. 
For example, the sample average in $\cals$ of $Y^*_i$ within a given leaf $\ell(x;\calp)$ will only be equal to $\hat\tau(x;\calp,\cals)$ if the fraction of treated observations within the leaf is exactly equal to $p$. % Effectively, taking the average of $Y^*_i$ as an estimator for the treatment effect within the leaf fails to adjust for the actual sample proportions.  % If the proportion treated in a leaf is higher than the overall sample, the treated observations are given more weight than the control observations when calculating the difference in sample means.   
Since this method is primarily considered as a benchmark, in simulations we focus only on an adaptive version that can use existing learning methods
entirely off-the-shelf.  The adaptive version of the transformed outcome tree estimator we consider, TOT-A,  uses the conventional CART algorithm with the transformed outcome replacing the original outcome.  The honest version, TOT-H, uses the same splitting and
cross-validation criteria, so that it builds the same trees; it differs only in that a separate estimation sample is used to construct
the leaf estimates.  The treatment effect estimator within a leaf is the same as the adaptive method, that is, the sample
mean of $Y_i^*$ within the leaf.
% Following standard CART, we use a penalty term that is linear in the number of leaves. The splitting is based on the standard criterion $\mse(\cals^{\train},\cals^{\train},\calp)$ discussed above, and the cross-validation is based on  $\mse(\cals^{\train,\tcv},\cals^{\train,\train},\calp)$. 

%\subsubsection{Honest Version}The honest version based on the transformed outcome changes the splitting criterion to $\widehat{\emse}(\cals^\train,\calp) $. The cross-validation criterion is changed to  $\widehat{\emse}(\cals^{\train,\tcv},\calp) $.

\subsection{Fit-based Trees (F)}

We consider two additional alternative methods for constructing trees, based on suggestions in the literature. 
In the first of these alternatives the choice of which feature to split on, and at what value of the feature to split, is based  on comparisons of the  goodness--of--fit of the outcome rather than the treatment effect. 
In standard CART of course goodness--of--fit of outcomes is also the split criterion, but here we estimate a model for treatment
effects within each leaf. Specifically, we have a linear model with an intercept and an indicator for the treatment as the regressors, rather only an intercept as in standard CART.  % Another way to look at it is that within each leaf we estimate outcomes separately for the treatment and control groups.  Our treatment effect estimate is the difference of those estimates, but for splitting and cross-validation, we focus on the fit of predictions of outcomes, where outcomes are predicted separately for treatment and control.
This approach is used in \cite{ZHH}, who consider building general models at the leaves of the trees.  Treatment
effect estimation is a special case of their framework.   \cite{ZHH} propose using statistical tests based on improvements in goodness-of-fit
to determine when to stop growing the tree, rather than relying on cross-validation, but for ease of comparison to CART, in this
paper we will stay closer to traditional CART in terms of growing deep trees and pruning them. %, instead only changing the  criteria.
%\[\hat\mu_w(W_i,X_i;\cals^\est,\calp)) = FILL IN \]
We modify the mean-squared error function:
\[ \msemw(\cals^\test,\cals^\est,\calp) \equiv \sum_{i\in\cals^\test} ((Y_i^\obs-\hat\mu_w(W_i,X_i;\cals^\est,\calp))^2-Y^2_i).\]
For the adaptive version F-A we follow conventional CART, using the criterion $- \msemw$ in place of $-\mse$ for splitting,
and the analog of $- \widehat{\mse}(\cals^{\train,\tcv},\cals^{\train,\train},\calp)$ with with $\hat\mu_w$ in place of $\hat\mu$ for cross-validation. For the honest version we use the analogs
of $-\widehat{\emse}(\cals^\train,\calp)$ and $-\widehat{\emse}(\cals^{\train,\tcv},\calp)$, with $\hat\mu_w$ in place of $\hat\mu$, for splitting and cross-validation. Similar to the prediction case, the variance term in the honest splitting criterion does not make much of a difference for the choice of splits.
 An advantage of the fit-based tree approach is that it is a straightforward extension of conventional CART methods.  In particular, the mean-squared error criterion is feasible, since $Y_i$ is observed.
To highlight the disadvantages of the F approach, consider a case where two splits improve the fit to an equal degree. In one case, the split leads to variation in average treatment effects, and in the other case it does not. The first split would be better from the perspective of estimating heterogeneous treatment effects, but the fit criterion would view the two splits as equally attractive.

%NEED TO AT LEAST DEFINE HONEST AND DISHONEST VERSION--CANNOT LEAVE THIS LIKE THIS

%For the adaptive version of fit-based trees, denoted F-A, we use  $ \widehat{\mse}_{\mu,w}(\cals^{\train},\cals^{\train},\calp)$ as the splitting criterion. This is essentially the same as standard CART, except that we replace $\hat\mu$ with $\hat\mu_w$. As in the regular CART algorithm the criterion will keep improving with each additional split. Following CART, to choose the complexity penalty we use cross-validation. The criterion is $ \widehat{\mse}_{\mu,w}(\cals^{\train,\tcv},\cals^{\train,\train},\calp)$. 

%For the honest version of fit-based trees, denoted F-H, we define $\widehat{\emse_{\mu,W}}(\cals^\train,\calp)$ by  substituting $\hat\mu_w$ for $\hat\mu$ in the definitions of $\widehat{\emse}(\cals^\train,\calp)$ and $S^2_{\cals^\train}(\ell).$ Then, we specify the splitting criterion to be $\widehat{\emse}_{\mu,w}(\cals^{\train},\calp)$, while for cross-validation we use $\widehat{\emse}_{\mu,w}(\cals^{\train,\tcv},\calp)$.

\subsection{Squared T-statistic Trees (TS)}

For the last estimator we look for splits with the largest value for the square of the
t-statistic for testing the null hypothesis that the average treatment effect is the same in the two potential leaves.
This estimator was proposed by \cite{STWNL}.
% They propose using a threshold for the T-statistic, where no splits are taken if the square of the t-statistic does not exceed athreshold, and thus the method does not rely on cross-validation.
 If the two leaves are denoted $L$ (Left)  and $R$ (Right), the square of the t-statistic is
\[ T^2 \equiv N\cdot \frac{(\oy_L-\oy_R)^2}{S^2/N_L+ S^2/N_R},\]
where $S^2$ is the conditional sample variance given the split.
At each leaf, successive splits are determined by selecting the split that maximizes $T^2$. 
The concern with this criterion is that it places no value on splits that improve the fit. While such splits do not deserve as much weight as the fit criterion puts on them, they do have some value.

%This criterion has two appealing features.  First, it potentially avoids the need for cross-validation.  However, this will potentially miss more complex interactions among covariates, so that the optimal cutoff may vary from application to application.  Thus, in this paper we implement a version augmented by cross-validation.  Second, this approach focuses on treatment effects, avoiding the shortcoming of the fit-based trees.  As we will discuss further below, this criterion actually goes too far:it  ignores some splits that improve the fit of the model, for example, cases where a covariate has a strong impact on outcomes, so that creating more homogenous leaves reduce the variance of outcomes within a leaf. Such splits improve the ability to estimate average treatment effects, even if there is no variation in the average treatment effects across leaves. 

Both the adaptive and honest versions of the TS approach use $T^2$ as the splitting criterion. For cross-validation and pruning, it is less
obvious how to proceed. 
%, because the standard CART uses goodness of fit measures that are defined at the individual unit level and thus can be summed across the children of a parent node; the goodness of fit measure is used to determine the ``weakest link'' to prune for a given penalty parameter (the SI Appendix  gives details).  To address this problem,
 \cite{ZHH} suggests that 
when using a statistical test for splitting, if it is desirable in an application to grow deep trees and then cross-validate to determine
depth, % (for example, if we hypothesize that there are important interactions among covariates where multiple splits are required to discover a strong effect), 
then one can use a standard goodness of fit measure for pruning and cross-validation.
%  This would suggest using the criteria from the fit-based trees, F-A and F-H, for adaptive and honest versions.
  However, this could
undermine the key advantage of TS, to focus on heterogeneous treatment effects.  For this reason, we instead propose
to use the CT-A and CT-H criteria for cross-validation for TS-A and TS-H, respectively. 

%For the adaptive version of cross validation TS-A we use  $ \widehat{\mse}_\tau(\cals^{\train,\tcv},\cals^{\train,\train},\calp)$. For the honest version of cross-validation TS-H we use $\widehat{\emse}_\tau(\cals^{\train,\tcv},\calp)$.

\subsection{Comparison of the Causal Trees, the Fit Criterion, and the Squared t-statistic Criterion}

It is useful to compare our proposed criterion to the F and TS criteria in a simple setting to gain insight into the relative merits of the three approaches. We do so here focusing on a decision whether to proceed with a single possible split, based on a binary covariate $X_i\in\{L,R\}$.
Let $\calp_N$ and $\calp_S$ denote the trees without and with the split, and
let $\oy_w$, $\oy_{Lw}$ and $\oy_{Rw}$ denote the average outcomes for units with treatment status $W_i=w$. Let $N_{w}$, $N_{Lw}$, and $N_{Rw}$ be the sample sizes for the corresponding subsamples.
Let $S^2$ be the sample variance of the outcomes given a split,
%\[ S^2=\frac{1}{N}\sum_{w=0}^1 \left( \sum_{i:W_i=w,X_i=L} (Y_i-\oy_{Lw})^2+\sum_{i:W_i=w,X_i=R} (Y_i-\oy_{Rw})^2 \right)\]
and let $\tilde S^2$ be the sample variance without a split.
%\[\tilde  S^2=\frac{1}{N}\sum_{w=0}^1  \sum_{i:W_i=w} (Y_i-\oy_{w})^2. \]
Define the squared t-statistics for testing that the average outcomes for control (treated) units in both leaves are identical,
\[T_0^2 \equiv \frac{(\oy_{L0}-\oy_{R0})^2}{S^2/N_{L0}+S^2/N_{R0}},\ \ 
 T_1^2 \equiv \frac{(\oy_{L1}-\oy_{R1})^2}{S^2/N_{L1}+S^2/N_{R1}}.\]
Then we can write the improvement in goodness of fit from splitting the single leaf into two leaves as
\[ F= \tilde S^2\cdot \frac{2\cdot (T_0^2+T_1^2)}{1+2\cdot (T_0^2+T_1^2)/N}.\]
Ignoring degrees-of-freedom correctcions, the change in our proposed criterion for the honest version of the causal tree in this simple setting can be written as a combination of the F and TS criteria:
\[ \widehat{\emse}_\tau(\cals,\calp_N)-\widehat{\emse}_\tau(\cals,\calp_S)=\frac{(T^2-4) (\tilde S^2-F/N)+2\tilde S^2 }{ p\cdot (1-p)}.\]
 Our criterion focuses primarily on $T^2$. Unlike the TS approach, however, it incorporates the benefits of splits due to improvement in the fit.

\section{Inference}

Given the estimated conditional average treatment effect we also would like to do inference. Once constructed, the tree is a 
function of covariates, and if we use a distinct sample to 
conduct inference, then the problem reduces to that of estimating treatment effects in each member of a partition of the covariate space.  For this problem, standard approaches are therefore valid for the estimates obtained via honest estimation, and in particular,
no assumptions about model complexity are required. For the adaptive methods standard approaches to confidence intervals are not generally valid for the reasons discussed above, and below we document through simulations that this can be important in practice.

% To be more precise, conditional on the tree $\calp$, consider the leaf $\ell$. Within this leaf the  average treatment effect is \[ \tau_{\ell}=\mathbb{E}[Y_i(1)-Y_i(0)|X_i\in\ell].\]Becauseof the randomization of units into training and estimation samples, we can view the the data for the subset of the estimation sample with features in this subset of the featurespace as arising from a completely randomized experiment. Hence the difference in average outcomes by treatment status is unbiased for the average effect in this subset of thefeature space, and we can estimate the variance without bias using the sample variance of the treated and control units in this subset.  

%To be specific, given a dataset, we propose to randomly select observations for a training sample $\cals^\train$ and estimation sample $\cals^\est$.  We can apply any partitioning algorithm $\pi$ to obtain $\pi(\cals^\train)$, so long as only data from the training sample is used in training and cross-validation.  Given this partition,  we estimate average treatment effects according to $\hat\tau(X_i;\cals^\est,\pi(\cals^\train)$, using  the estimation sample $\cals^\est$. Then conditional on the training sample, the variance of the treatment effect estimate in leaf $\ell$ is $S^{2}_{\cals^\est_\treat}(\ell)/N^\est_{\treat}(\ell)+S^{2}_{\cals^\est_\control}(\ell)/N^\est_{\control}(\ell)$.  For each leaf, confidence intervals can then be constructed around the treatment effect estimate.

\section{A Simulation Study}

To assess the relative performance of the proposed algorithms we carried out a small simulation study with three distinct designs.  In Table 1 we report a number of
summary statistics from the simulations. We report averages; results for medians are similar.  We report results for 
$N^\train=N^\est$ with either 500 or 1000 observations.  When comparing adaptive to honest approaches, we report the ratio of the $\mset$ for adaptive estimation with $N^\train=1000$ to $\mset$ for honest estimation
with $N^\train=N^\est=500$, in order to highlight the tradeoff between sample
size and bias reduction that arises with honest estimation.  We evaluate $\mset$ using a test sample with $N^\test=6000$ observations to test the methods in order to minimize the sampling variance in our simulation results.

In all designs, the marginal treatment probability is $p= 0.5$. $K$ denotes the number of features. In each design, we have a model $\eta(x)$ for the mean
effect and $\kappa^\text{sim}(x)$ for the treatment effect.  Then, the potential outcomes are written 
\[ Y_i(w) = \eta^\text{sim}(X_i) + \frac{1}{2} \cdot (2w-1) \cdot \kappa(X_i) + \epsilon_i ,\]
where $\epsilon_i \sim {\cal N}(0,.01)$, and the $X_i$ are independent of $\epsilon_i$ and one another, and $X_i \sim {\cal N}(0,1)$.
The designs are summarized as follows:
\[\text{1: } K=2; \eta(x)=\frac{1}{2}x_1+ x_2; \kappa(x)=\frac{1}{2}x_1. \]
\[\text{2: } K=10; \eta(x)=\frac{1}{2}\sum_{k=1}^2 x_k+\sum_{k=3}^6 x_k; \kappa(x)=\sum_{k=1}^2 1\{x_k>0\} \cdot x_k \]
\[\text{3: } K=20; \eta(x)=\frac{1}{2}\sum_{k=1}^4 x_k+\sum_{k=5}^8 x_k; \kappa(x)=\sum_{k=1}^4 1\{x_k>0\} \cdot x_k \]
In each design, there are some covariates that affect treatment effects ($\kappa$) and mean outcomes ($\eta$); some covariates
that enter $\eta$ but not $\kappa$; and some covariates that do not
affect outcomes at all (``noise'' covariates).  Design 1 does not have noise covariates.  In Designs 2 and 3, the first few covariates
enter $\kappa$, but only when their signs are positive, while they affect $\eta$ throughout their range.  Different criterion will thus
lead to different optimal splits, even within a covariate; F will focus more on splits when the covariates are negative.

The first panel of Table 1 compares the number of leaves in different designs and different values of $N^\train=N^\est$.  Recalling that TOT-A and TOT-H have the same splitting method, we see that it tends to build shallow trees.  The failure to control for the 
realized value of $W_i$ leads to additional noise in estimates, which tends to lead to aggressive
pruning.  For the other estimators, the adaptive versions lead to shallower trees than the honest versions, as the honest versions
correct for overfitting, and the main cost of small leaf size is high variance in leaf estimates.  F-A and F-H are very similar; as discussed above, the splitting criterion are very similar, and further,
the F estimators are less prone to overfitting treatment effects, because they split based upon overall model fit.  We also observe
that the F estimators build the deepest trees; they reward splitting on covariates that affect mean outcomes as well as treatment effects.

The second panel of Table 1 examines the performance of the alternative honest estimators, as evaluated by the infeasible
 criterion $\mset$.  We report the average of the ratio of $\mset$ for a given estimator to $\mset$ for our preferred
estimtor, CT-H.  The TOT-H estimator performs well in Designs 2 and 3, but suffers in Design 1.  In Design 1, the variance of
$Y_i$ conditional on $(W_i,X_i)$ is very low at $.01$, and so the failure of TOT to account for the realization of $W_i$ results
in a noticeable loss of performance. The F-H estimator suffers in all 3 designs; all designs give the F-H criterion attractive
opportunities to split based on covariates that do not enter $\kappa$.  F-H would perform better in alternative designs
where $\eta(x)=\kappa(x)$; F-H also does well at avoiding splits on noise covariates.  The TS-H estimator performs 
well in Design 1, where $x_1$ affects $\eta$ and $\kappa$ the same way, so that the CT-H criterion is aligned with TS-H.
Design 3 is more complex, and the ideal splits from the perspective of balancing overall mean-squared error of treatment effects
(including variance reduction) are different from those favored by TS-H.  Thus, TS performs worse, and the difference is
exacerbated with larger sample size, where there are more opportunities for the estimators to build deeper trees and thus to make different choices.    We also calculate comparisons based on a feasible criterion, the average squared difference between the transformed outcome $Y^*_i$ and the estimated treatment effect $\hat\tau_i$. For details for this comparison see the SI Appendix. In general the results are consistent with those from the infeasible criterion.

The third panel of Table 1 explores the costs and benefits to honest estimation.  The Table reports the ratio of $\mse_\tau(\cals^\test,\cals^\est \cup \cals^\train,\pi^\text{Estimator-A}(\cals^\est \cup \cals^\train))$ to $\mse_\tau(\cals^\test,\cals^\est,\pi^\text{Estimator-H}(\cals^\train)$ for each estimator.  The adaptive version uses the union of the training and estimation
samples for tree-building, cross-validation, and leaf estimation.  Thus it has double the sample size (1000 observations) at each step,
while the honest version uses 500 of the observations in training and cross-validation, with the complement used for estimating
treatment effects within leaves.  The results
show that there is a cost to honest estimation in terms of $\mset$, varying by design and estimator.
%, ranging from : the honest version has higher mean-squared error of treatmenteffects.  The cost varies from design to design; forCT-H, it is less than 10\% in Designs 1 and 2, but larger in Design 3.  Design 3has a large amount of treatment effectheterogeneity, and so the additional data allows more complex trees, and the gains to complexity are very large.

The final two panels of Table 1 show the coverage rate for 90\% confidence intervals.  We achieve nominal coverage rates for honest
methods in all designs, where, in contrast,  %; this is expected, but it is worth highlighting that with 20 covariates, 500 observations, and a rich model of interactions, we would not expect nominal coverage rates for traditional nonparametric methods that do not rely %on partitions. In thischallenging environment, our method produces an average of 6.2 subgroups and discovers meaningful %treatment effectheterogeneity with confidence intervals. In
 the adaptive methods have coverage rates substantially below nominal rates. %, due to the confidence intervals not being centered on the true treatment effects.  
Thus, our simulations bear out the tradeoff that honest estimation sacrifices some goodness of fit (of
treatment effects) in exchange for valid confidence intervals.

\section{Observational Studies with Unconfoundedness}\label{SecUnconf}

The discussion so far has focused on the setting where the assignment to treatment is randomized. The proposed methods can be adapted to observational studies under the assumption of unconfoundedness. In that case we need to modify the estimates within leaves to remove the bias from simple comparisons of treated and control units. There is a large literature on methods for doing so, e.g., \cite{IR15}. For example, as in \cite{HIR} we can do so by propensity score weighting.
% The simplest way to implement this is to weight each observation by the inverse of a normalized propensity score and apply the algorithms using the weighted outcomes asthe outcomevariable;
Efficiency will improve if we renormalize the weights within each leaf and and within the treatment and control group when estimating treatment effects. \cite{CHIM} propose approaches to trimming observations with extreme values for the propensity score to 
improve robustnesses. 
% This ensures that treatment and control units receive equal weight when taking the difference of sample averages. 
 Note that there are some additional conditions required to establish asymptotic normality 
of treatment effect estimates when propensity score weighting is used (see, e.g., \cite{HIR}); these results apply without
modification to the estimation phase of honest partitioning algorithms.

%For inference, outside of the setting of simple randomized assignment, under the unconfoundedness assumption, we can estimate treatment effects using inverse propensity score methods.  \cite{HIR} establish conditions under which the estimated treatment effects are  asymptotically normal.  For those conditions to be satisfied in this application, it is necessary to restrict the size of the leaves of the tree relative to the size of the sample.  One could also use other methods for estimating treatment effects within the leaves on the test sample, such as matching methods.  Matching is computationally costly, which is particularly problematic during training, but may be less of a concern for a single applica]tion in the test sample. 

\section{The Literature}

A small but growing literature seeks to apply supervised machine learning techniques to the problem of estimating heterogeneous treatment effects.  Beyond those previously discussed,
\cite{TAGT} transform the features rather than the outcomes and then apply LASSO to the model with the original outcome and the transformed features.
% This requires the model to be linear in the features and their interactions with the treatment so that it does not directly extend to tree methods.
\cite{FTR} estimate $\mu(w,x)=\mme[Y_i(w)|X_i=x]$  for $w=0,1$ using random forests, then calculate $\hat\tau_i=\hat\mu(1,X_i)-\hat\mu(0,X_i)$.  They then use machine learning algorithms to estimate $\hat\tau_i$ as a function of the units' attributes, $X_i$.  Our approach differs in that we apply machine learning methods directly to the treatment effect in a single stage procedure.
\cite{IR13} %consider multi-valued treatments but focus on the purely experimental setting.  They 
 use LASSO to estimate the effects of both treatments  and attributes, but with different penalty terms for the two types of features to allow for the possibility that the treatment effects are present but the magnitudes of the interactions are small.  Their approach is similar to ours in that they distinguish between the estimation of treatment effects and the estimation of the impact of other
attributes of units.
\cite{TGCD} consider a model with the outcome linear in the covariates and the interaction with the treatment variable. Using Bayesian nonparametric methods with Dirichlet priors, they project their estimates of heterogeneous treatment effects down onto the feature space using LASSO-type regularization methods to get low-dimensional summaries of the heterogeneity. 
%\cite{ZHH} develop a procedure they call ``model-based recursive partitioning'' whereby they develop a tree-based method for estimating parametric models on subsets of the data.  At each leaf of the tree, they propose to estimate a model such as a linear regression or a maximum-likelihood based model.  Leaves are split further using a test for parameter stability; thefeature with the highest instability is chosen for a split.  In terms of a method for building a tree, this approach is similar to the causal tree in that we estimate a simple model (estimatetreatment effects) within leaves of the tree, and we split the leaves when we find covariates that lead to different parameter estimates within the splits (heterogeneous treatmenteffects).  \cite{ZHH} differ in that they base the split in sample on model fit, and in that they do not consider cross-validation for selecting a complexity tuning parameter, so theissue of selecting an out-of-sample goodness of fit metric does not arise.    
%\cite{STWNL} construct a tree with the splitting  based on the t-statistic for the test of no difference between the two groups. The cross-validation of the overall tree is based on the sum of the squares of the split t-statistics, with a penalty term that is linear in the number of splits. Our splitting criteria is conceptually similar, but our approach considers alternative loss functions for cross-validation that directly assess goodness of fit of estimated treatment effects.
%We discussed \cite{BL} above.  NO WE DIDN''T THAT WAS CUT 
\cite{DLL} and \cite{BL} propose a related appoach for finding the optimal treatment policy
 that combines inverse propensity score methods with ``direct methods'' (e.g. the ``single tree'' approach considered above) that predict the outcome as a function of the treatment and the unit attributes.  The methods can be
used to evaluate the average difference in outcomes from any two policies that map attributes to treatments, as well as to select the optimal policy function. They do not focus
on hypothesis testing for heterogeneous treatment effects, and they use conventional approaches for cross-validation. 
Also  related is the work on Targeted Learning \cite{VR}, which
modifies the loss function to increase the weight on the parts of the likelihood that concern the parameters of interest. 
% The methods rely on a parametric model for the parameters of interest whe; in contrast, we consider nonparametric estimates of $\tau(x)$ in an environment where there may be a large number of covariates $x$ relative to the sample size.

\section{Conclusion} 

In this paper we introduce new methods for constructing trees for causal effects that allow us to do valid inference for the causal effects in randomized experiments and in observational studies satisfying unconfoundedness, without restrictions on the number of covariates or the complexity of the data generating process. Our methods partition the feature space into subspaces.  The output of our method is a set of treatment effects and confidence intervals for each subspace.  

A potentially important application of the techniques is to ``data-mining'' in randomized experiments.  Our method can be used
to explore any previously conducted randomized controlled trial, for example, medical studies or field experiments in developed economics.  A researcher can apply our methods and discover subpopulations with lower-than-average or higher-than-average treatment
effects, and can report confidence intervals for these estimates without concern about multiple testing.

\newpage 
\begin{table}[ht]
\centering
%\caption{Table 1}
\label{my-label}
\begin{tabular}{lccccccc}
\multicolumn{7}{c}{Table 1: Simulation Study}\\
\multicolumn{7}{c}{}\\
Design &  \multicolumn{2}{c}{1} & \multicolumn{2}{c}{2} & \multicolumn{2}{c}{3} \\
 $N^\train=N^\est$ & 500 & 1000 & 500 & 1000 & 500 & 1000 \\
Estimator &  \multicolumn{6}{c}{Number of Leaves} \\
TOT               & 2.8        & 3.4        & 2.1        & 2.7        & 4.7       & 6.1       \\
F-A               & 6.1        & 13.2       & 6.3        & 13.1       & 6.1       & 13.2      \\
TS-A              & 4.0        & 5.6        & 2.5        & 3.3        & 4.4       & 8.9       \\
CT-A              & 4.0        & 5.7        & 2.3        & 2.5        & 4.5       & 6.2       \\
F-H               & 6.1        & 13.2       & 6.4        & 13.3       & 6.3       & 13.4      \\
TS-H              & 4.4        & 7.7        & 5.3        & 11.0       & 6.0       & 12.3      \\
CT-H              & 4.2        & 7.5        & 5.3        & 11.2       & 6.2       & 12.3      \\
                   \multicolumn{7}{c}{Infeasible MSE Divided by Infeasible MSE for CT-H$^*$} \\
TOT-H             & 1.77       & 2.12       & 1.03       & 1.04       & 1.03      & 1.05      \\
F-A               & 1.93       & 1.54       & 1.69       & 2.07       & 1.63      & 2.08      \\
TS-H              & 1.01       & 1.02       & 1.06       & 0.99       & 1.24      & 1.38      \\
CT-H              & 1.00       & 1.00       & 1.00       & 1.00       & 1.00      & 1.00      \\
                   \multicolumn{7}{c}{Ratio of Infeasible MSE: Honest to Adaptive$^{**}$}    \\
TOT-H/TOT-A       &            & 0.99       &            & 0.86       &           & 0.76      \\
F-H/F-A           &            & 0.50       &            & 0.98       &           & 0.91      \\
TS-H/TS-A         &            & 0.92       &            & 0.90       &           & 0.85      \\
CT-H/CT-A         &            & 0.91       &            & 0.93       &           & 0.76      \\
                   \multicolumn{7}{c}{Coverage of 90\% Confidence Intervals - Adaptive}      \\
TOT-A             & 0.83       & 0.86       & 0.83       & 0.83       & 0.74      & 0.79      \\
F-A               & 0.89       & 0.89       & 0.86       & 0.86       & 0.82      & 0.82      \\
TS-A              & 0.85       & 0.85       & 0.80       & 0.83       & 0.77      & 0.80      \\
CT-A              & 0.85       & 0.85       & 0.81       & 0.83       & 0.80      & 0.81      \\
                   \multicolumn{7}{c}{Coverage of 90\% Confidence Intervals - Honest}        \\
TOT-H             & 0.90       & 0.89       & 0.90       & 0.92       & 0.89      & 0.89      \\
F-H               & 0.91       & 0.90       & 0.90       & 0.90       & 0.90      & 0.89      \\
TS-H              & 0.89       & 0.90       & 0.90       & 0.90       & 0.90      & 0.90      \\
CT-H              & 0.90       & 0.90       & 0.90       & 0.89       & 0.90      & 0.90     
\end{tabular}
{\small $^*\mse_\tau(\cals^\test,\cals^\est,\pi^\text{Estimator}(\cals^\train))/\mse_\tau(\cals^\test,\cals^\est,\pi^\text{CT-H}(\cals^\train))$

$^{**}\mse_\tau(\cals^\test,\cals^\est \cup \cals^\train,\pi^\text{Estimator-A}(\cals^\est \cup \cals^\train))/$
\hskip 4mm $\mse_\tau(\cals^\test,\cals^\est,\pi^\text{Estimator-H}(\cals^\train)$}
\end{table}

 \newpage

\centerline{\large Additional Simulation Details (Online Appendix)}
\vskip 5mm

This Appendix describes some additional details of the simulation study, and also presents additional simulation results in Appendix 
Table A1.

The code for our simulations was written as an R software package that is in preparation for public release.  It is based on the `rpart'
R package, available at https://cran.r-project.org/web/packages/rpart/index.html. For TOT, we directly use rpart
applied to the transformed outcome $Y_i^*$, and we use 10-fold cross-validation for pruning the tree.    For each of our other estimators, we modified several components of 
the package.  In the remainder of this Appendix, discussions of modifications apply to F, CT, and TS estimators.  For these estimators, we create new versions of the ``anova'' functions, functions that in the standard rpart package are used for calculating the total goodness of fit for a node of the tree,
evaluating the quality of alternative splits, and estimating the goodness of fit for pruned trees using cross-validation samples.  We maintain the overall
structure of the rpart package.  The rpart package has an important tuning parameter, which is the minimum number of observations
per leaf, denoted $n_m$.  We modify the rpart routine to insist on at least $n_m$ treated {\it and} $n_m$ control observations per leaf, to ensure
that we can calculate a treatment effect within each leaf.  In the simulations reported in Table 1 of the paper and in this Appendix,
we use $n_m=25$ for all models except TOT, while for the TOT model the minimum leaf size is 50 (without restrictions on treated
and control observations). 

We make one additional modification to the way the standard rpart splitting function works.  We restrict the set of potential split points considered,
and further, in the splitting process we rescale the covariate values within each leaf and each treatment group
in order to ensure that when moving
from one potential split point to the next, we move the same number of treatment and control observations from the right leaf to the left leaf.  We begin by describing the motivation
for these modifications, and then we give details.

The rpart algorithm considers every value of $X_{i,k}$ in $\cals^\train$ as a possible split point for covariate $X_k$.
An obvious disadvantage of this approach is that computation time can grow prohibitively large in models with many observations
and covariates.  But there are some more subtle disadvantages as well. The first is that there will naturally be sampling variation
in estimates of $\hat\tau$ as we vary the possible split points. A problem akin to a multiple hypothesis testing problem arises:
since we are looking for the maximum value of an estimated criterion across a large number of possible split points,
as the number of split points tested grows, it becomes more and more likely that one of the splits for a given covariate appears to improve the fit criterion even if the true
value of the criterion would indicate that it is better not to split.  One way to mitigate both the computation time problem and
the multiple-testing problem is to consider only a limited number of split points. 

A third problem is specific to considering treatment effect heterogeneity.  To see the problem, suppose that a covariate
strongly affects the mean of outcomes, but not treatment effects.  Within a leaf, some observations are treated and some
are control. If we consider every level of the covariate in the leaf as a possible split point, then shifting from one split point to the
next shifts a single observation from the right leaf to the left leaf.  This observation is in the treatment or the control group,
but not both; suppose it is in the treatment group.  If the covariate has a strong effect on the level of outcomes, the observation that is shifted will be likely have an
outcome more extreme than average. It will change the sample average of the treatment group, but not the control group, leading
to a large change in the estimated treatment effect difference.
We expect the estimated difference in treatment effects across the left and right leaves to fluctuate greatly with the split point in this scenario.  This
variability around the true mean difference in treatment effects occurs more often when covariates affect mean outcomes, and thus it
can lead the estimators to split too much on such covariates, and also to find spurious opportunities to split.

To address this problem, we propose the following modifications to the splitting rule.  We include a parameter $b$, 
the target number of observations per ``bucket.'' For each leaf, before testing possible
splits for a particular covariate, we order the observations by the covariate value in the treatment and control group separately.  Within each group, we place the observations into buckets with $b$ observations per bucket.  If this results in less than $n_m$ buckets, then we use fewer observations
per bucket (to attain $n_m$ buckets). We number the buckets, and considering splitting by bucket number rather than the raw
values of the covariates.  This guarantees that when we shift from one split point to the next, we add both treatment and
control observations, leading to a smoother estimate of the goodness of fit function as a function of the split point.  After the
best bucket number to split on is selected, we translate that into a split point by averaging the maximum covariate value in
the corresponding treatment and control buckets.  In the simulations presented in this paper, we do not constrain the maximum number of buckets, and we let $b=4$.  We found that this discretization approach improved goodness of fit on average for the 
simulations we considered, although
it can in principle make things worse.

In the simulations reported in Table 1 of ths paper, we used the infeasible $\mset$ to evalute alternative estimators.  In practice,
we must estimate the infeasible criterion. In the paper, we propose estimators that rely on the tree structure of our estimator,
but we may also wish to compare our performance to estimators that don't rely on partitions.  One alternative
is the $MSE^\text{TOT}$ criterion.  Given an estimator $\hat\tau_i$, it is equal to
\[MSE^\text{TOT}= \frac{1}{N^\test}\sum_{i=1}^{N^\test} (Y^*_i-\hat\tau_i)^2.\]
Because $\mme[Y_i^*|X_i]=\tau(X_i)$, this is an unbiased (but noisy) estimator for $\mset$. In Appendix Table A1, we present rankings
of estimators using this criterion. We see that it ranks estimators in the same way as $\mset$ except in one case (Design 3 with 500 observations), but the percentage differences between
estimators are smaller than with the infeasible criterion.

Another tuning parameter for standard CART as well as the methods proposed here is the number of cross-validation samples.  A 
common convention is to use 10 samples. We deviate from that convention and use 5 cross-validation samples.  The reason is
that our methods require various quantities to be estimated within leaves.  Given a minimum leaf restriction of 25 treated and control
units, if we take a cross-validation sample of one-tenth of the original training sample, we might end up with no treated or no
control observations in a leaf in a cross-validation sample. In addition, it may be difficult to estimate a sample variance within a leaf.
Rather than require larger leaf sizes, we simply use fewer cross-validation samples.  

Appendix Table A1 also includes the full set of estimates for the infeasible criterion $\mset$, to illustrate how sample size and
honest versus adaptive estimation affects the criterion.  Note that for purposes of comparison to the simulation results from Table 1,
Table 1 reports the average over simulations of the ratio of the goodness of fit measures; the second panel of this table shows
the average of goodness of fit measures, but the ratio of the averages shown here is not exactly equal to the average of the ratios
shown in Table 1.

\newpage

\begin{table}[ht]
\centering
Appendix Table A1: Infeasible and Feasible MSE Estimates for Simulation Study
\begin{tabular}{lcccccc}
\multicolumn{7}{c}{}\\
Design &  \multicolumn{2}{c}{1} & \multicolumn{2}{c}{2} & \multicolumn{2}{c}{3} \\
 $N^\train=N^\est$ & 500 & 1000 & 500 & 1000 & 500 & 1000 \\
Estimator &  \multicolumn{6}{c}{$MSE^\text{TOT}_\tau$ Divided by $MSE^\text{TOT}_\tau$ for CT-H$^*$} \\
TOT-H  & 1.010  & 1.009  & 1.001  & 1.001  & 1.004  & 1.008 \\
F-H  & 1.013  & 1.004  & 1.039  & 1.049  & 1.121  & 1.161 \\
TS-H   & 0.999  & 1.000  & 1.003  & 0.999  & 1.046  & 1.056 \\
CT-H   & 1.000  & 1.000  & 1.000  & 1.000  & 1.000  & 1.000 \\
\multicolumn{7}{c}{Infeasible $\mset$}                            \\
TOT-A  & 0.171  & 0.139  & 1.267  & 1.010  & 3.235  & 2.344 \\
F-A    & 0.150  & 0.075  & 1.914  & 1.861  & 4.914  & 4.443 \\
TS-A   & 0.103  & 0.077  & 1.509  & 1.080  & 4.068  & 3.169 \\
CT-A   & 0.104  & 0.079  & 1.418  & 1.071  & 3.324  & 2.314 \\
TOT-H  & 0.140  & 0.104  & 1.176  & 0.932  & 3.102  & 2.252 \\
F-H    & 0.151  & 0.075  & 1.898  & 1.850  & 4.860  & 4.420 \\
TS-H   & 0.083  & 0.053  & 1.201  & 0.887  & 3.732  & 2.935 \\
CT-H   & 0.087  & 0.054  & 1.149  & 0.910  & 3.033  & 2.143 \\
\end{tabular}
{\small $^*\mse^\text{TOT}_\tau(\cals^\test,\cals^\est,\pi^\text{Estimator}(\cals^\train))/\mse^\text{TOT}_\tau(\cals^\test,\cals^\est,\pi^\text{CT-H}(\cals^\train))$}
\end{table}

\end{document}